\documentclass{llncs}
\usepackage{url}
\usepackage{amssymb}

\usepackage{soul}

\usepackage[within=none]{newfloat}
\DeclareFloatingEnvironment[name={Supplementary Figure}]{suppfigure}

\usepackage[colorlinks]{hyperref}
\usepackage[colorinlistoftodos]{todonotes}
\usepackage{verbatim}
\usepackage{subfig}
\usepackage{booktabs}
\usepackage{graphicx}

\begin{document}

\title{Graph Saliency Maps through Spectral Convolutional Networks: Application to Sex Classification with Brain Connectivity}
%
%
 \author{Salim Arslan, Sofia Ira Ktena, Ben Glocker, Daniel Rueckert}
 \authorrunning{S. Arslan et al.} 
 \tocauthor{Salim Arslan, Sofia Ira Ktena, Ben Glocker, Daniel Rueckert}
 \institute{Biomedical Image Analysis Group, Department of Computing,\\
  Imperial College London, UK}

\maketitle              

\begin{abstract}
Graph convolutional networks (GCNs) allow to apply traditional convolution operations in non-Euclidean domains, where data are commonly modelled as irregular graphs. Medical imaging and, in particular, neuroscience studies often rely on such graph representations, with brain connectivity networks being a characteristic example, while ultimately seeking the locus of phenotypic or disease-related differences in the brain. These regions of interest (ROIs) are, then, considered to be closely associated with function and/or behaviour. Driven by this, we explore GCNs for the task of ROI identification and propose a visual attribution method based on class activation mapping. By undertaking a sex classification task as proof of concept, we show that this method can be used to identify salient nodes (brain regions) without prior node labels. Based on experiments conducted on neuroimaging data of more than 5000 participants from UK Biobank, we demonstrate the robustness of the proposed method in highlighting reproducible regions across individuals. We further evaluate the neurobiological relevance of the identified regions based on evidence from large-scale UK Biobank studies.

\end{abstract}

\section{Introduction} 
Graph convolutional neural networks (GCNs) have recently gained a lot of attention, as they allow adapting traditional convolution operations from Euclidean to irregular domains~\cite{bronstein2017geometric}. Irregular graphs are encountered very often in medical imaging and neuroscience studies in the form of brain connectivity networks, supervoxels or meshes. In these cases, applications might entail both node-centric tasks, e.g. node classification, as well as graph-centric tasks, e.g. graph classification or regression. While CNNs have redefined the state-of-the-art in numerous problems by achieving top performance in diverse computer vision and pattern recognition tasks, insights into their underlying decision mechanisms and the impact of the latter on performance are still limited.

Recent works in deep learning address the problem of identifying salient regions in 2D/3D images in order to visualise determinant patterns for classification/regression tasks performed by a CNN and obtain spatial information that might be useful for the delineation of regions of interest (ROI)~\cite{simonyan2013deep}. In the field of neuroscience, in particular, the identification of the exact locus of disease- or phenotype-related differences in the brain is commonly sought. Locating brain areas with a critical role in human behaviour and mapping functions to brain regions as well as diseases on disruptions to specific structural connections are among the most important goals in the study of the human connectome.

In this work, we explore GCNs for the task of brain ROI identification. As proof of concept, we undertake a sex classification task on functional connectivity networks, since there is previous evidence for sex-related differences in brain connectivity~\cite{satterthwaite2014linked}. Characteristically, stronger functional connectivity was established within the default mode network of female brains, while stronger functional connectivity was found within the sensorimotor and visual cortices of male brains~\cite{Ritchie123729}. As a result, we consider this a suitable application to demonstrate the potential of the proposed method for delineating brain regions based on the attention/sensitivity of the model to the sex of the input subject's connectivity graph. More specifically, we show that spatially segregated salient regions can be identified in non-Euclidean space by using class activation mapping~\cite{zhou2016learning} on GCNs, making it possible to effectively map the most important brain regions for the task under consideration.

\textbf{Related work:}
Graph convolutions have been employed to address both graph-centric and node-centric problems and can be performed in the spatial~\cite{monti2017geometric} or spectral domain~\cite{defferrard2016convolutional,levie2017cayleynets}. In the latter case, convolutions correspond to multiplications in the graph spectral domain and localised filters can be obtained with Chebyshev polynomials~\cite{defferrard2016convolutional} or rational complex functions~\cite{levie2017cayleynets}.~\cite{zhou2018convolution} introduced adaptive graph convolutions and attention mechanisms for graph- and node-centric tasks, while in~\cite{velivckovic2017graph} attention mechanisms were employed to assign different weights to neighbours in node classification tasks with inductive and transductive inference. Although the latter works focus the attention of the network onto the most relevant nodes, they overlook the importance/contribution of different features/graph elements for the task at hand.

At the same time, visual feature attribution through CNNs has attracted attention, as it allows identifying salient regions in an input image that lead a classification network to a certain prediction. It is typically addressed with gradient and/or activation-based strategies. The former relies on the gradients of the prediction with respect to the input and attributes saliency to the regions that have the highest impact on the output~\cite{simonyan2013deep}. Activation-based methods, on the other hand, associate feature maps acquired in the final convolutional layer with particular classes and use weighted activations of the feature maps to identify salient regions~\cite{zhou2016learning,selvaraju2016grad}. A recent work addresses the problem from an adversarial point of view and proposes a visual attribution technique based on Wasserstein generative adversarial networks~\cite{baumgartner2017visual}. While these methods offer promising results on Euclidean images, their application to graph-structured data is yet to be explored. 

\textbf{Contributions:} We propose a visual feature attribution method for graph-structured data by combining spectral convolutional networks and class activation mapping~\cite{zhou2016learning}. Through a graph classification task, in which each graph represents a brain connectivity network, we detect and visualise brain regions that are responsible for the prediction of the classifier, hence providing a new means of brain ROI identification. As a proof of concept, we derive experiments in the context of sex differences in functional connectivity. First, we  train a spectral convolutional network classifier and achieve state-of-the-art accuracy in the prediction of female and male subjects based on their functional connectivity networks captured at rest. The activations of the feature maps are, then, used for visual attribution of the nodes, each of which is associated with a brain region. Using resting-state fMRI (rs-fMRI) data of more than 5000 subjects acquired by UK Biobank, we show that the proposed method is highly robust in selecting the same set of brain regions/nodes across subjects and yields highly reproducible results across multiple runs with different seeds.

\section{Method}
\begin{figure}[tbh]
	\centering
    \includegraphics[width=\textwidth]{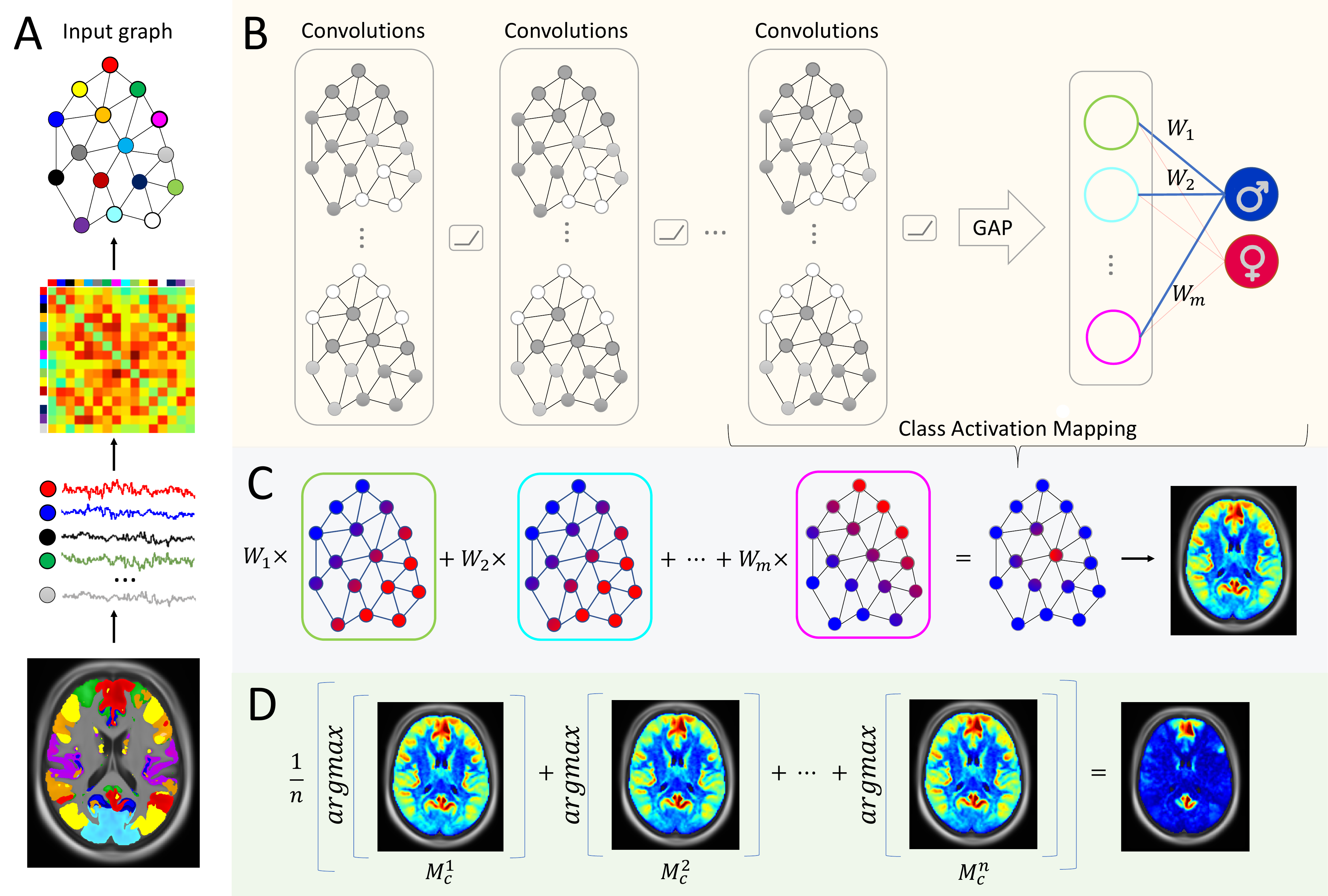} 
	\caption{Overview of the proposed approach. (A) The input graph is computed using a brain parcellation and rs-fMRI connectivity signals. (B) Graph convolutional network model. Convolutional feature maps of the last layer are spatially pooled via global average pooling (GAP) and connected to a linear sex classifier. (C) Class activation mapping procedure. (D) Generation of population-level saliency maps.} 
	\label{fig:pipelines}
    \vspace{-0.8cm}
\end{figure}

Fig.~\ref{fig:pipelines} illustrates the proposed method for identifying brain regions used by GCNs to predict a subject's sex based on its functional connectivity. Given an adjacency matrix that encodes similarities between nodes and a feature matrix representing a node's connectivity profile, the proposed method outputs the sex of the input subject and provides a graph saliency map highlighting the brain regions/nodes that lead to the corresponding prediction. Finally, we rank brain regions with respect to their contribution towards driving the model's prediction at subject level and compute a population-level saliency map by combining them across individuals.

\subsubsection{Spectral graph convolutions:}
We assume $n$ samples (\emph{i.e.} subjects), $X = [X_1, \ldots, X_n]^T$, with signals defined on a graph structure. Each subject is associated with a data matrix $X_{i} \in \mathbb{R}^{d_x \times d_y}$, where $d_y$ is the dimensionality of the node's feature vector, and a label $y_i \in \{0, 1\}$. In order to encode the structure of the data, we define a weighted graph $G = (V, E, W)$ where $V$ is the set of $d_x = |V|$ nodes (vertices), $E$ is the set of edges (connections) and $W \in \mathbb{R}^{d_x \times d_x}$ is the weighted adjacency matrix, representing the weight of each edge, \emph{i.e.} $W_{i,j}$ is the weight of the edge connecting $v_i \in V$ to $v_j \in V$. 

A convolution in the graph spatial domain corresponds to a multiplication in the graph spectral domain. Hence, graph filtering operations can be performed in the spectral domain using the eigenfunctions of the normalised Laplacian of a graph~\cite{shuman2013emerging}, which is defined as $L=I_{d_x} - D^{-\frac{1}{2}}WD^{-\frac{1}{2}}$, where $D$ is the degree matrix and $I_{d_x}$ the identity matrix. In order to yield filters that are strictly localised and efficiently computed, Defferrard et al.~\cite{defferrard2016convolutional} suggested a polynomial parametrisation on the Laplacian matrix by means of Chebyshev polynomials. Chebyshev polynomials are recursively computed using $T_k(L)= 2LT_{k-1}(L)-T_{k-2}$, with $T_0(L)=1$ and $T_1(L)=L$.

A polynomial of order $K$ yields strictly $K$-localised filters. Filtering of a signal $x$ with a $K$-localised filter can, then, be performed using:

\begin{equation}
y = g_\theta(L) \ast x = \sum_{k=0}^{K} \theta_k T_k(\tilde{L})x,
\end{equation}

with $\tilde{L}=\frac{2}{\lambda_{max}}L-I_{d_x}$ and $\lambda_{max}$ denoting the largest eigenvalue of the normalised Laplacian, $L$. The output of the $l^{th}$ layer for a sample $s$ in a graph convolutional network is, then, given by:

\begin{equation}
y_s^l=\sum_{i=1}^{F_{in}} g_{\theta_i^l}(L)x^l_{s,i}.
\end{equation}

For $F_{out}$ output filter banks and $F_{in}$ input filter banks, this yields $F_{in} \times F_{out}$ vectors of trainable Chebyshev coefficients $\theta_i^l \in \mathbb{R}^K$ with $x^l_{s,i}$ denoting the input feature map $i$ for sample $s$ at layer $l$. Hence, at each layer the total number of trainable parameters is $F_{in} \times F_{out} \times K$.

\subsubsection{Class activation mapping:} Class activation mapping (CAM)~\cite{zhou2016learning} is a technique used to identify salient regions that assist a CNN to predict a particular class. It builds on the fact that, even though no supervision is provided on the object locations, feature maps in various layers of CNNs still provide reliable localisation information~\cite{zhou2014object}, which can be captured via global average pooling (GAP) in the final convolutional layer~\cite{lin2013network}. Encoded into a class activation map, these ``spatially-averaged'' deep features not only yield approximate locations of objects, but also provide information about where the attention of the model focuses when predicting a particular class~\cite{zhou2016learning}. In the context of GCNs, CAM is used to localise discriminative nodes, each associated with a saliency score.

The process for generating class activation maps is illustrated in Fig~\ref{fig:pipelines}. Given a typical GCN model which consists of a series of convolutional layers, a GAP layer is inserted into the network right after the last convolutional layer. The spatially-pooled feature maps are connected to a dense layer that produces the output for a classification task (Fig.~\ref{fig:pipelines}B). We can then linearly map the weights of the dense layer onto the corresponding feature maps to generate a class activation map showing the salient nodes in the graph (Fig.~\ref{fig:pipelines}C).

More formally, let $f_i(v)$ represent the activation of the $i$th feature map in the last convolutional layer at node $v$. For the feature map $i$, the average pooling operation is defined as $F_i=(1/d_z)\sum_{v}^{}f_i(v)$, where $F_i \in \mathbb{R}$ and $d_z$ is the number of nodes in the feature map. Thus, for a given class $c$, the input to the dense layer is $\sum_{i}w_{i}^{c}F_i$, where $w_{i}^{c}$ is the corresponding weight of $F_i$ for class $c$. Intuitively, $w_{i}^{c}$ indicates the importance of $F_i$ for class $c$, therefore, we can use these weights to compute a class activation map $M_c$, where each node is represented by a weighted linear sum of activations, \emph{i.e.} $M_c(v)=\sum_{i}w_{i}^{c}f_i(v)$. This map shows the impact of a node $v$ to the prediction made by the GCN model and, once projected back onto the brain, can be used to identify the ROIs that are most relevant for the specific classification task. 

\subsubsection{Population-level saliency maps:} Although CAM provides graph-based activation maps at subject/class-level, population-level statistics about discriminative brain regions are also important. In order to combine class activation maps across subjects, we define a simple $argmax$ operation that, for each subject, returns the index of the $k$ top nodes with the highest activation. These are, subsequently, averaged across subjects and referenced as the population-level saliency maps as illustrated in Fig~\ref{fig:pipelines}D.

\subsubsection{Network architecture and training:} The details of the GCN architecture are presented in Table~\ref{tab:model} and summarised as follows. 5 convolutional layers, each succeeded by rectified linear (ReLU) non-linearity are used. No pooling is performed between consecutive layers, as empirical results suggest that reducing the resolution of the underlying graph does not improve performance. We apply zero-padding to keep the spatial resolution of the feature maps unchanged throughout the model. A dropout rate of 0.5 is used in the $2^{nd}$, $4^{th}$, and $5^{th}$ layers. The feature maps of the last layer are spatially averaged and connected to a linear classifier with softmax output. We employ global average pooling as it reflects where the attention of the network is focused and substantially reduces the number of parameters, hence alleviating over-fitting issues~\cite{lin2013network}. 

\begin{table}[tb]
\centering
\caption{Network architecture of the proposed model. * indicates the use of dropout for the corresponding convolutional layer.}
\begin{tabular}{lcccccccc}
\toprule
\textbf{Layer}&Input&Conv&Conv*&Conv&Conv*&Conv*&GAP&Linear\\
\midrule
\textbf{Channels}&55&32&32&64&64&128&128&2\\
\textbf{\emph{K}-order }&N/A&9&9&9&9&9&N/A&N/A\\
\textbf{Stride}&N/A&1&1&1&1&1&N/A&N/A\\
\bottomrule
\end{tabular}
\label{tab:model}
\end{table}

The loss function used to train the model comprises a cross entropy term and an $L_2$ regularisation term with decay rate of $5e^{-4}$. We use an Adam optimiser with momentum parameters $\beta = [0.9, 0.999]$ and initialise the training with a learning rate of 0.001. Training is performed for a fixed number of 500 steps (\emph{i.e.} 20 epochs), in mini-batches of 200 samples, equally representing each class. We evaluate the model every 10 steps with an independent validation set, which is also used to monitor training. Based on this, the learning rate is decayed by a factor of 0.5, whenever validation accuracy drops in two consecutive evaluation rounds.

\section{Data and Experiments}
\subsubsection{Dataset and preprocessing:} Imaging data is collected as part of UK Biobank's health imaging study (\url{http://www.ukbiobank.ac.uk/}), which aims to acquire imaging data for 100,000 predominantly healthy subjects. The multimodal scans together with the vast amount of non-imaging data are publicly available to assist researchers investigating a wide range of diseases, such as dementia, arthritis, cancer, and stroke. We conduct our experiments on rs-fMRI images available for 5430 subjects from the initial data release. Non-imaging data and medical information are also provided alongside brain scans including sex, age, genetic data, and many others. The dataset used here consists of 2873 female (aged 40-70 yo, mean $55.38 \pm 7.41$) and 2557 male (aged 40-70 yo, mean $56.61 \pm 7.60$) subjects. 

Details of data acquisition and preprocessing procedures are given in~\cite{miller2016multimodal}. Standard preprocessing steps have been applied to rs-fMRI images including motion correction, 
high-pass temporal filtering, and gradient distortion correction. 
An independent component analysis (ICA)-based approach is used to identify and remove structural artefacts~\cite{miller2016multimodal}. Finally, images go through visual quality control and any preprocessing failures are eliminated.

\subsubsection{Brain parcellation and network modelling:} A dimensionality reduction procedure known as ``group-PCA''~\cite{smith2014group} is applied to the preprocessed data to obtain a group-average representation. This is fed to group-ICA~\cite{beckmann2004probabilistic} to parcellate the brain into 100 spatially independent, non-contiguous components. Group-ICA components identified as artefactual (\emph{i.e.} not neuronally-driven) are discarded and the remaining $d=55$ components are used to estimate functional connectivity for each subject by means of $L_2$-regularised partial correlation between the ICA components' representative timeseries~\cite{smith2011network}.

\subsubsection{Experimental setup:} We use stratified 10-fold cross-validation to evaluate the model with split ratios set to 0.8, 0.1, and 0.1 for training, validation, and testing, respectively. Cross-validation allows to use all subjects for both training/validation and testing, while each subject in the dataset is used for testing exactly once. To further evaluate how the performance varies across different sets of subjects and how robust the identified salient regions are, we repeat cross-validation 10 times with different seeds.

\section{Results and Discussion}
In Table~\ref{tab:acc} we provide classification results obtained with the GCN classifier. The presented accuracy rates correspond to the results of all 10 folds for each run. On average, we achieve a test accuracy of $88.06\%$ across all runs/folds, with low standard deviation for each run, indicating reproducible classification performance. It is worth noting that, although classification is not the end goal of the proposed method, a high accuracy rate is a prerequisite for robust and reliable activation maps. Yet, the average performance of our classifier is slightly higher than the state-of-the-art sex classification accuracy with respect to functional connectivity~\cite{ktena2018,arslan2018}. 

\begin{table}[b]
\centering
\caption{Average sex classification accuracy rates (in \%) for each run.}
\begin{tabular}{lccccccccccc}
\toprule
\textbf{Run}&~1~&~2~&~3~&~4~&~5~&~6~&~7~&~8~&~9~&~10~&Avr\\
\midrule
\textbf{Acc}&~88.51~&~88.27~&~87.64~&~87.94~&~87.84~&~88.01~&~88.51~&~88.08~&~88.05~&~87.77~&~\textbf{88.06}\\
\textbf{Std}&1.57&1.25&1.88&1.30&1.73&1.66&1.54&1.34&1.93&1.06&\textbf{1.57}\\
\bottomrule
\end{tabular}
\label{tab:acc}
\end{table}

\begin{figure}[t]
	\centering
    \includegraphics[width=\textwidth]{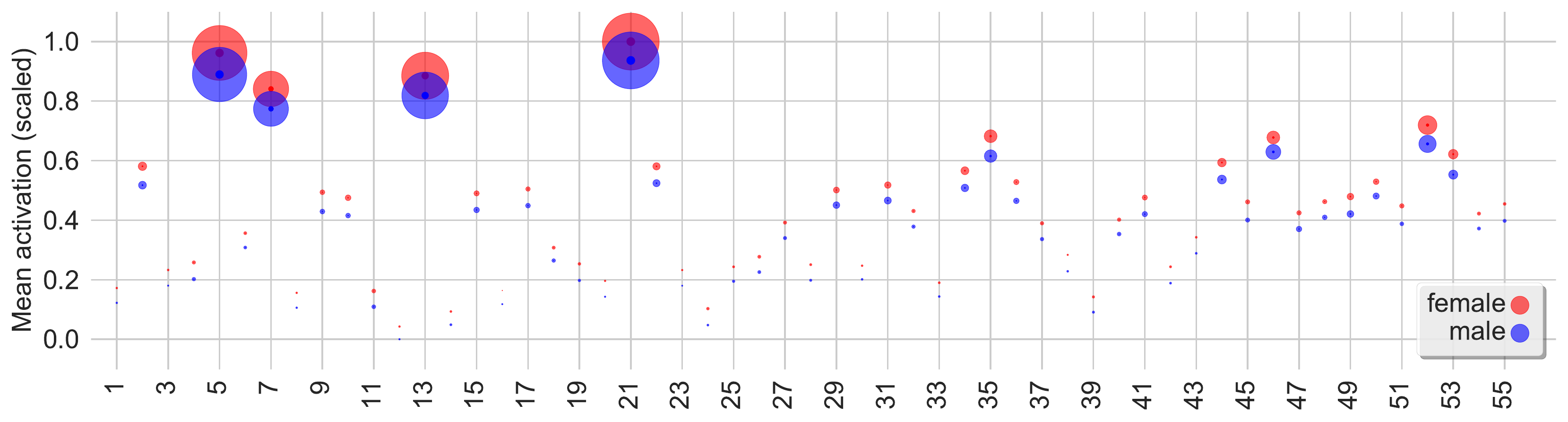} 
	\caption{Sex-specific class activations for all nodes averaged across subjects and runs. Mean activations are scaled to $[0,1]$ for better visualisation. The size of the markers indicates the number of times a node is ranked within the top $k=3$ most important, summed across subjects and runs. } 
	\label{fig:results}
    \vspace{-0.5cm}
\end{figure}

Fig.~\ref{fig:results} shows the sex-specific activations for all nodes. As illustrated, the GCN focuses on the same regions for both classes, with one class (\emph{i.e.} female) consistently yielding higher activation than the other (\emph{i.e.} male). This can be attributed to the fact that a binary classifier only needs to predict one class, while every other sample is automatically assigned the remaining class label. The most important nodes, in descending order, are 21, 5, 13, and 7. As indicated by the size of their markers, these four nodes are almost always ranked within the top $k=3$ of all nodes with respect to their activations, meaning that all subjects but few are consistently classified according to the connections of these nodes. While we only provide results for $k=3$, the same regions are identified for lower/higher values of $k$, with only minimal changes in their occurrence rate, as shown in Supplementary Fig.~\ref{sup:results}. 

\begin{figure}[thb]
	\centering
    \includegraphics[width=\textwidth]{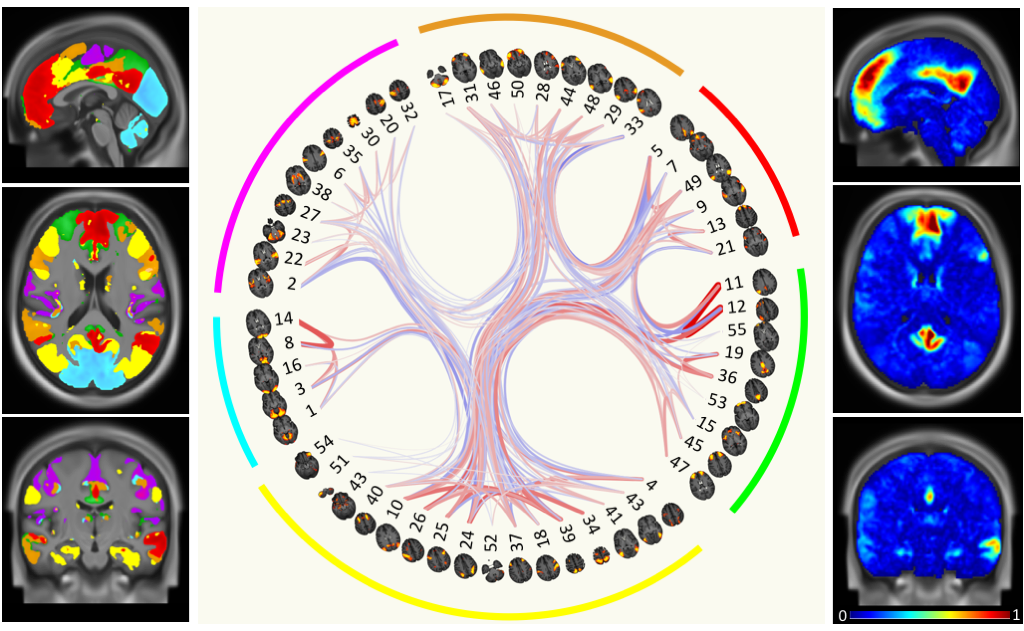} \hfill
	\caption{\emph{Left}: ICA-based brain parcellation shown in groups of six resting-state networks (RSNs), including the default mode network (red). The tree slices shown are, from top to bottom, sagittal, axial, and coronal, at indices 91, 112, and 91, respectively. \textit{Middle}: Connectogram showing the group-averaged functional connectivity between 55 brain regions, which are clustered based on their average population connectivity. Strongest positive and negative correlations are shown in red and blue, respectively. Image is adapted from \url{http://www.fmrib.ox.ac.uk/ukbiobank/netjs_d100/} and enhanced for better visualisation \emph{Right}: Population-level saliency maps, combined for both genders.} 
	\label{fig:connectome}
    \vspace{-0.5cm}
\end{figure}

In order to explore the neurobiological relevance of these results, we refer to the UK Biobank group-averaged functional connectome~\cite{miller2016multimodal}, which maps the functional interactions between the 55 brain regions clustered into six resting state networks (RSNs) according to their average population connectivity (Fig.~\ref{fig:connectome}). RSNs comprise spatially segregated, but functionally connected cortical regions, that are associated with diverse functions, such as sensory/motor, visual processing, auditory processing, and memory. Our comparisons to the connectome revealed that regions 21, 5, 13, and 7 (as shown in Fig.~\ref{fig:connectome}) are part of the default mode network (highlighted with red), a spatially distributed cluster which is activated `by default' during rest. A large-scale study on sex differences in the human brain~\cite{Ritchie123729} has also found evidence that functional connectivity is stronger for females in the default mode network, which might further indicate that the identified regions are neurobiologically relevant and reflect sex-specific characteristics encoded in functional connectivity.

\section{Conclusion}
In this paper, we have addressed the visual attribution problem in graph-structured data and proposed an activation-based approach to identify salient graph nodes using spectral convolutional neural networks. By undertaking a graph-centric classification task, 
we showed that a GCN model enhanced with class activation mapping can be used to identify graph nodes (brain regions), even in the absence of supervision/labels at the node level. Based on experiments conducted on neuroimaging data from UK Biobank, we demonstrated the robustness of the proposed method by means of highlighting the same regions across different subjects/runs using cross validation. We further validated the neurobiological relevance of the identified ROIs based on evidence from UK Biobank studies~\cite{miller2016multimodal,Ritchie123729}.

While the potential of the proposed method is demonstrated on functional networks, it can be applied to any graph-structured data. However, its applicability might be limited by several factors, including the definition and number of nodes (\emph{e.g.} brain parcellation), network modelling, as well as node signal choices. It is also important to assess the robustness of the identified regions by disentangling the effect of the graph structure and the node features. Future work will focus on experiments in this direction, as well as the applicability of the method to other graph-centric problems (\emph{e.g.} regression). For instance, a GCN model can be trained for age prediction and consequently used to identify brain regions for which connectivity is most affected with aging. Another interesting direction entails extending this work for directed/dynamic, \emph{e.g.} time-varying, graphs.

\subsubsection*{Acknowledgements.} 
This research has been conducted using the UK Biobank Resource under Application Number 12579 and funded by the EPSRC Doctoral Prize Fellowship funding scheme.

\bibliographystyle{splncs}
\bibliography{bibliography}

\begin{suppfigure}[bth]
	\centering
    \includegraphics[width=\textwidth]{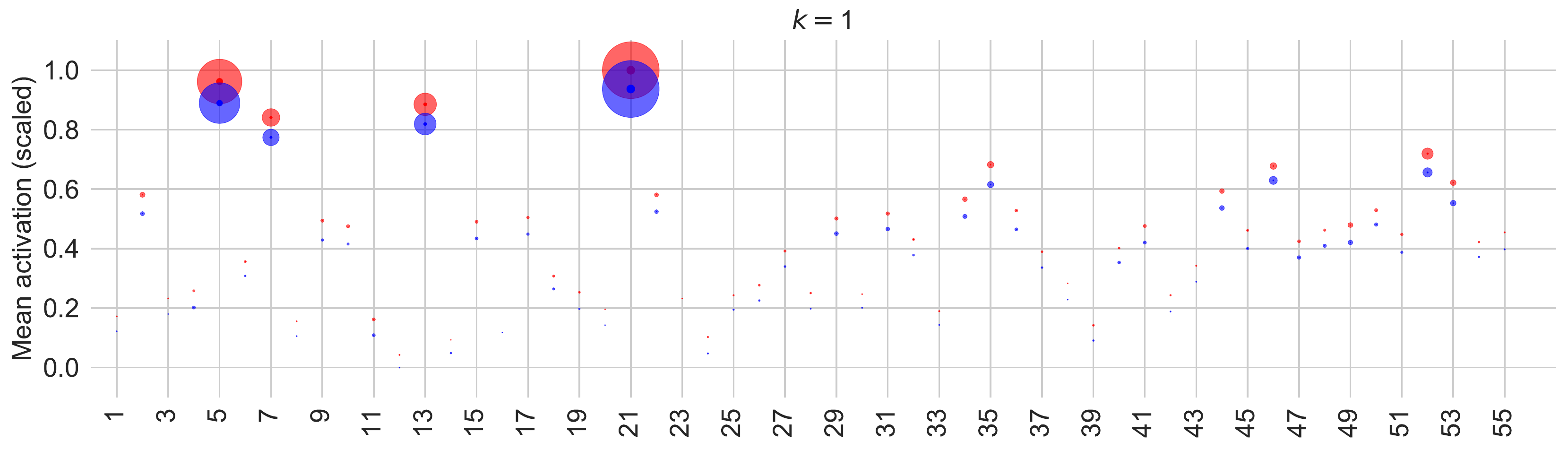} \\ \hfill
    \includegraphics[width=\textwidth]{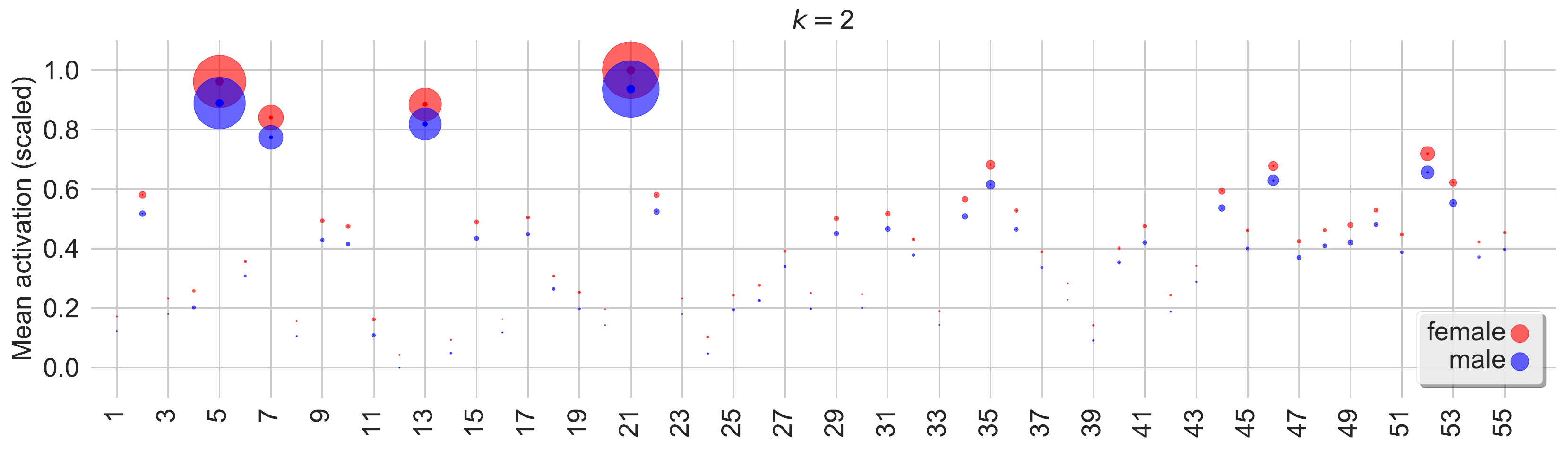} \\ \hfill
    \includegraphics[width=\textwidth]{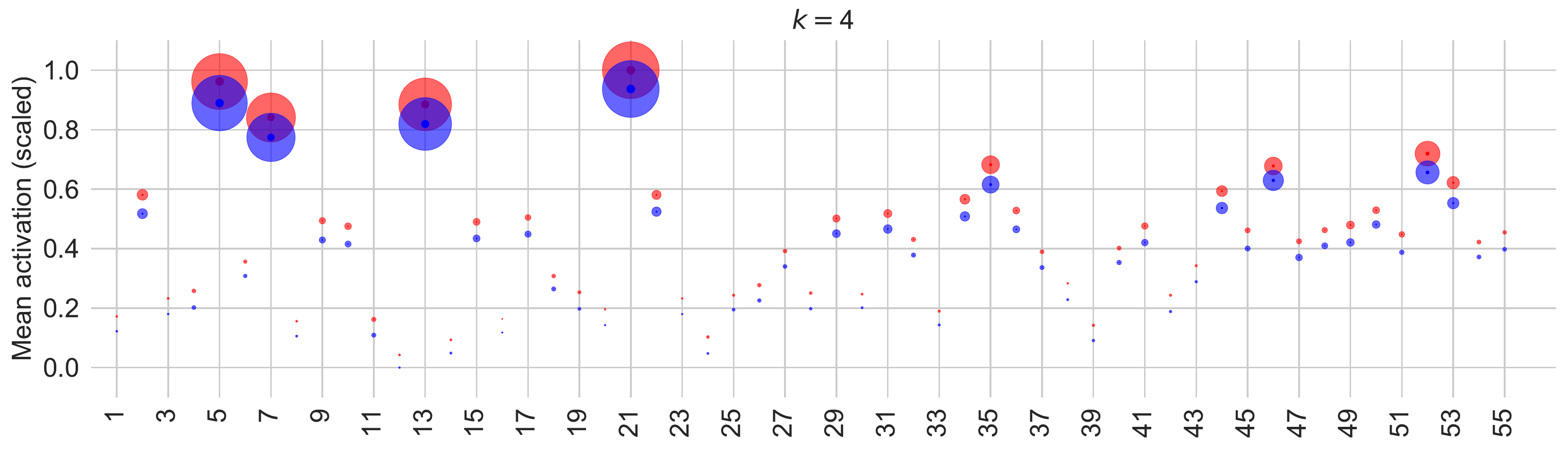} \\
	\caption{Sex-specific class activations for all nodes averaged across subjects and runs. Mean activations are scaled to $[0,1]$ for better visualisation. Size of a marker indicates the number of times a node is ranked within the top $k=1, 2, 4$ of all nodes, summed across subjects and runs. } 
	\label{sup:results}
\end{suppfigure}

\end{document}